# Handwritten Digit Recognition: An Ensemble-Based Approach for Superior Performance

Syed Sajid Ullah[1*]  · Li Gang[1], Mudassir Riaz[1], Ahsan Ashfaq[1], ,Salman Khan[1],Sajawal Khan[1]

[1] *Chang'an University, China*

\* Corresponding author.  Email: *sajid@chd.edu.cn*

| KEYWORDS | Handwritten digit recognition remains a fundamental challenge in computer vision, with applications ranging from postal code reading to document digitization. This paper presents an ensemble-based approach that combines Convolutional Neural Networks (CNNs) with traditional machine learning techniques to improve recognition accuracy and robustness. We evaluate our method on the MNIST dataset, comprising 70,000 handwritten digit images. Our hybrid model, which uses CNNs for feature extraction and Support Vector Machines (SVMs) for classification, achieves an accuracy of 99.30%. We also explore the effectiveness of data augmentation and various ensemble techniques in enhancing model performance. Our results demonstrate that this approach not only achieves high accuracy but also shows improved generalization across diverse handwriting styles. The findings contribute to the development of more reliable handwritten digit recognition systems and highlight the potential of combining deep learning with traditional machine learning methods in pattern recognition tasks. |
|---|---|
| Handwritten Digit Recognition; Convolutional Neural Networks; Support Vector Machines; Ensemble Learning; Hybrid Models; | |

## 1. Introduction

Handwritten digit recognition is a fundamental challenge in the fields of machine learning and computer vision, with significant applications in automated data entry, postal mail sorting, and bank check processing. As a benchmark problem, it serves as a reference for evaluating the performance of various classification algorithms, due to its practical relevance and the inherent complexities arising from diverse handwriting styles. The MNIST dataset, consisting of 70,000 images of handwritten digits, has become a standard resource for assessing algorithmic performance in this domain. It provides a rich, representative dataset that captures the variability of human writing [1], [2].

Over the years, handwritten digit recognition methodologies have evolved from traditional machine learning techniques, such as k-nearest neighbors and support vector machines (SVMs), to more advanced deep learning architectures, particularly Convolutional Neural Networks (CNNs). These models have demonstrated exceptional performance in recognizing handwritten digits by automatically learning hierarchical feature representations from pixel data [3]. Notably, CNN architectures, such as LeNet-5, have set new benchmarks in accuracy, achieving recognition rates exceeding 99% on the MNIST dataset. This highlights the transformative impact of deep learning in this field [3].

Despite these advancements, challenges remain in handwritten digit recognition, especially in dealing with variations in writing styles, digit overlap, and image noise. To address these issues, researchers have increasingly adopted hybrid approaches that combine the strengths of both traditional machine learning and deep learning techniques. For example, data augmentation and transfer learning have proven to be effective strategies for improving model robustness and generalization [4].

In this paper, we propose an ensemble-based approach that combines Convolutional Neural Networks (CNNs) with traditional machine learning techniques, specifically Support Vector Machines (SVMs), to address the challenges of handwritten digit recognition. Our approach leverages CNNs for effective feature extraction and SVMs for accurate classification, achieving an impressive recognition accuracy of 99.30% on the MNIST dataset. Additionally, we explore the impact of data augmentation and ensemble techniques on enhancing model performance and generalization across diverse handwriting styles. The findings of this work not only contribute to improving the reliability of handwritten digit recognition systems but also demonstrate the potential of integrating deep learning with traditional machine learning methods to solve complex pattern recognition tasks.

**2. Related Work**

Handwritten digit recognition has been a focal point of research in the fields of machine learning and computer vision, evolving significantly over the past few decades. Early approaches primarily utilized traditional machine learning techniques, such as template matching and support vector machines (SVMs), which laid the groundwork for more complex models. However, the advent of deep learning has revolutionized this domain, particularly with the introduction of Convolutional Neural Networks (CNNs), which have demonstrated superior performance in recognizing handwritten digits. For instance, Memon et al. provide a comprehensive review of optical character recognition (OCR) systems, highlighting the shift towards deep learning methodologies that leverage advancements in computational power and neural network architectures [5]. This transition has enabled researchers to achieve unprecedented accuracy levels in digit recognition tasks.

Recent studies have further explored the integration of hybrid models that combine traditional machine learning techniques with deep learning approaches. For example, Putri emphasizes the importance of feature extraction in improving the performance of handwritten digit recognition systems [6]. The combination of CNNs with other classifiers, such as SVMs, has been shown to enhance recognition accuracy and robustness against variations in handwriting styles and noise . This hybrid approach not only improves accuracy but also facilitates the model's adaptability to diverse datasets, as demonstrated by Putri, who utilized incremental SVMs to refine recognition systems [6].

Moreover, the application of advanced data augmentation techniques has been pivotal in enhancing model performance. Research by Gao indicates that employing various augmentation strategies can significantly improve the generalization capabilities of CNNs, thereby addressing challenges posed by the inherent variability in handwritten digits [7].

The exploration of novel architectures has also gained traction in recent years. For instance, the work of Agrawal and Jagtap introduces a Convolutional Vision Transformer model that merges the strengths of CNNs and transformers, aiming to further enhance recognition performance on benchmark datasets [8]. This innovative approach reflects a broader trend in the field towards hybrid architectures that leverage the advantages of multiple neural network types to tackle complex recognition tasks.

Additionally, the use of ensemble methods has been highlighted as a promising strategy for improving classification outcomes. Research by Tuba et al. provides a survey of various classification methods tested on the MNIST dataset, noting that ensemble techniques can significantly enhance classification performance [9]. This aligns with the findings of Abdulhussain et al., who demonstrated that hybrid models utilizing orthogonal polynomials and moments achieved robust performance in noisy environments, outperforming traditional CNNs [10].

In summary, the landscape of handwritten digit recognition is characterized by a dynamic interplay between traditional machine learning methods and cutting-edge deep learning techniques. The integration of hybrid models, advanced data augmentation, and novel architectures continues to push the boundaries of accuracy and robustness in this field. As researchers strive to address the challenges posed by diverse handwriting styles and environmental noise, the ongoing evolution of methodologies promises to enhance the reliability and applicability of handwritten digit recognition systems across various domains.

**3. Methodology**

A structured approach was employed in this research, encompassing data preparation, neural network design, training strategies, and model evaluation techniques. The process involved preprocessing the MNIST dataset, designing and training deep learning models, and assessing their performance in handwritten digit recognition. Each step is outlined to provide a comprehensive understanding of the methods used to achieve optimal recognition results.

## 3.1. Dataset and Data Preparation

The MNIST dataset, a well-established benchmark for handwritten digit recognition, is employed in this study. This dataset consists of 70,000 grayscale images of handwritten digits, each represented as a 28x28 pixel matrix. Out of these, 60,000 images are reserved for training the models, while the remaining 10,000 are used for testing and validation purposes. The digits range from 0 to 9, and each image is labeled with the corresponding digit. The MNIST dataset is notable for its uniformity, with the digits centered in each image and normalized for consistent scale and position, ensuring that the data is ready for machine learning tasks without requiring extensive preprocessing.

Given the diversity in handwriting styles present in the dataset, additional preprocessing steps were applied to enhance model performance. Each image was normalized by scaling pixel values to a range between 0 and 1, which aids in faster convergence during training by standardizing the input data. Data augmentation techniques such as random rotations, shifts, and zooms were also implemented to artificially expand the training set and introduce variability. This helps in making the models more robust to slight changes in digit appearance, improving generalization when applied to unseen data.

The dataset is sourced from the United States Census Bureau and other institutional collections, ensuring that it represents a wide variety of handwriting styles. This diversity in data allows models to better generalize across different handwriting forms, making MNIST an ideal choice for evaluating deep learning models aimed at digit classification.

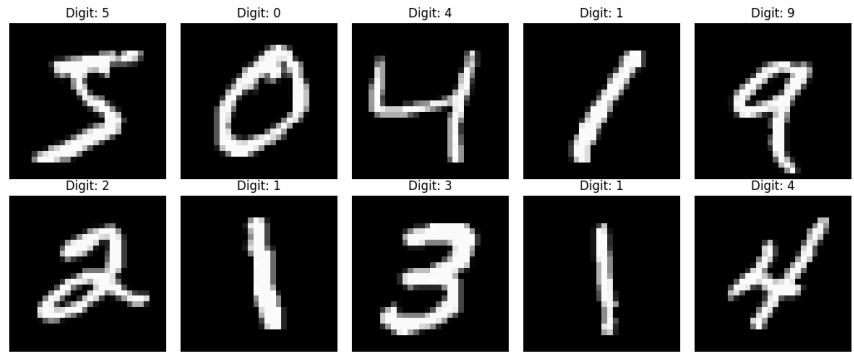

**Figure 1.** *Sample Images from MNIST Dataset*

### 3.1.1. Class Distribution

The class distribution of the MNIST dataset is uniform across all ten digits (0-9), ensuring balanced representation in both training and testing sets. This uniform distribution simplifies model training by preventing bias toward any particular class. However, despite this uniformity, the variations in handwriting styles across different samples introduce a level of complexity that necessitates the use of powerful deep learning models.

As shown in Figure 2, the distribution of digits in the MNIST dataset is balanced, with an approximately equal number of samples for each digit class. This ensures that models trained on this dataset do not favor one class over another, providing a fair evaluation of classification performance.

The data preparation phase involves several critical steps to ensure the dataset is suitable for analysis. Initially, the dataset is loaded and partitioned into training, validation, and test sets. This division is crucial for effective model evaluation. Pixel values are normalized to a consistent range between 0 and 1, enhancing the model's learning efficiency. Images are reshaped and scaled to standardize input dimensions, ensuring uniformity across the neural network.

Preprocessing techniques are applied to enhance the model's robustness and generalization capabilities. These techniques include pixel value normalization and data augmentation methods. Data augmentation involves transformations such as rotation, scaling, and translation to increase the variability of the training data. These augmentations help the model generalize better to new, unseen data and improve overall performance.

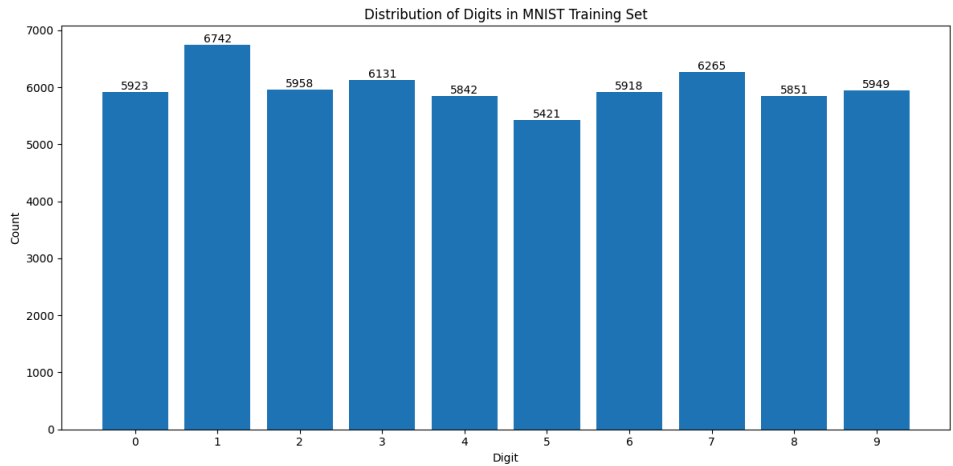

**Figure 2.** *Distribution of digits in MNIST dataset*

## 3.2. Model Architectures

This section presents an overview of the machine learning and deep learning architectures employed in the study. The models are classified into four primary categories: traditional machine learning models, ensemble and boosting methods, deep learning models, and hybrid models. Each category represents a distinct methodology for addressing complex classification challenges, with its associated strengths and limitations.

### 3.2.1. Traditional Machine Learning Models

Traditional machine learning models form the foundation of our comparative analysis. These models, while often simpler than their deep learning counterparts, offer interpretability and efficiency, making them valuable in many practical applications.

Logistic Regression, a fundamental technique in statistical modeling, serves as our baseline model. It operates by estimating the probability of an instance belonging to a particular class, utilizing a logistic function to map any real-valued number to a value between 0 and 1. Despite its simplicity, logistic regression remains a powerful tool, particularly in scenarios where linear separation is possible or when interpretability is crucial [11], [12].

Support Vector Machines (SVMs) represent a more sophisticated approach within the traditional machine learning paradigm. SVMs construct a hyperplane or set of hyperplanes in a high-dimensional space, aiming to maximize the margin between classes. This margin maximization contributes to SVMs' robustness and generalization capability [13], [14]. In non-linearly separable cases, SVMs employ kernel tricks to implicitly map the input space to a higher-dimensional feature space, enabling complex decision boundaries [15].

The k-Nearest Neighbors (k-NN) algorithm offers a non-parametric approach to classification. Unlike models that learn a fixed set of parameters, k-NN makes predictions based on the majority class among the k nearest neighbors in the feature space. This locality-based decision-making allows k-NN to capture complex patterns, albeit at the cost of increased computational complexity during inference [16].

Random Forest, an ensemble method, leverages the power of multiple decision trees. By constructing a multitude of decision trees at training time and outputting the mode of the classes as the final prediction, Random Forests mitigate the overfitting tendencies of individual decision trees. This approach often results in improved generalization and robustness to noise, making Random Forests a popular choice in various domains [7], [17].

Ensemble and boosting methods represent a significant advancement in machine learning, combining multiple models to produce superior predictive performance. In our study, we focus on XGBoost (eXtreme Gradient Boosting) as a representative of this category. XGBoost, an implementation of gradient-boosted decision trees, is designed for speed and performance. It builds trees sequentially, with each new tree aiming to correct the errors of the ensemble of existing trees. XGBoost incorporates several technical advancements, including a novel tree-learning algorithm for handling sparse data and a theoretically justified weighted quantile sketch procedure for approximate learning [18]. These innovations allow XGBoost to push the limits of computational resources for boosted tree algorithms, often resulting in state-of-the-art performance on a wide array of machine learning tasks [19], [20].

*3.2.2. Deep Learning Models*

Deep learning models have revolutionized the field of machine learning, particularly in domains involving large-scale, high-dimensional data. Our study includes three prominent deep learning architectures: Convolutional Neural Networks (CNNs), Recurrent Neural Networks (RNNs), and Long Short-Term Memory networks (LSTMs).

Convolutional Neural Networks have emerged as the architecture of choice for processing grid-like data, particularly images. CNNs leverage the principles of local connectivity and weight sharing through convolutional layers, enabling them to automatically and adaptively learn spatial hierarchies of features [21]. This architectural bias towards spatial invariance makes CNNs particularly effective in tasks such as image classification, object detection, and semantic segmentation [22].

Recurrent Neural Networks are designed to process sequential data, making them well-suited for tasks involving time series, text, or any data with inherent temporal or sequential structure. RNNs maintain an internal state (memory) that gets updated as they process input sequences, allowing them to capture temporal dependencies [23].However, vanilla RNNs often struggle with long-term dependencies due to the vanishing gradient problem [24].

Long Short-Term Memory networks, a specialized form of RNNs, address the limitations of vanilla RNNs in capturing long-term dependencies. LSTMs introduce a more complex unit with gates that regulate the flow of information, allowing the network to selectively remember or forget information over long sequences [25], [26]. This architecture has proven particularly effective in tasks such as machine translation, speech recognition, and complex sequence modeling [27].

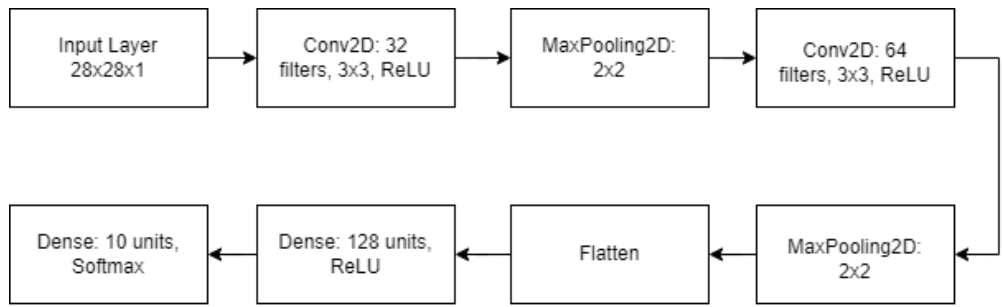

**Figure 3.** *Architectural design of CNN*

*3.2.3. Hybrid Models*

Hybrid models represent an emerging trend in machine learning research, combining different architectural approaches to leverage their respective strengths. Our study explores two such hybrid models: CNN-SVM and CNN-LSTM.

The CNN-SVM hybrid model utilizes a CNN for feature extraction followed by an SVM for classification. This architecture aims to combine the powerful feature learning capabilities of CNNs with the effective classification of SVMs. By using the CNN as a feature extractor, the model can automatically learn relevant features from raw input data, which are then fed into the SVM for final classification. This approach has shown promise in various image classification tasks, particularly when the amount of labeled data is limited.

The CNN-LSTM architecture combines the spatial feature extraction capabilities of CNNs with the sequential learning power of LSTMs. This hybrid model is particularly suited for tasks that involve both spatial and temporal components, such as video classification or image captioning. The CNN component processes spatial information from input data, while the LSTM component models the temporal dynamics or sequential relationships. This synergistic combination allows the model to capture both spatial and temporal dependencies effectively. These hybrid models represent an active area of research, showcasing the potential of combining diverse architectural paradigms to create more powerful and versatile machine learning systems.

### *3.3. Training and Validation*

The training process for the neural network involves using backpropagation and gradient descent optimization techniques. During training, the network weights are adjusted through backpropagation to minimize the loss function. Optimization algorithms such as Adam or Stochastic Gradient Descent (SGD) are used for iterative weight updates.

Key parameters in the training process include the learning rate, batch size, and the number of epochs. Regularization techniques, such as dropout, are applied to prevent overfitting. Training is performed on GPUs to accelerate computations.

Validation techniques are employed to monitor the model's performance and adjust hyperparameters. A separate validation set is used to fine-tune hyperparameters and ensure model generalization. Cross-validation methods may also be applied for additional reliability.

Validation metrics such as accuracy, precision, recall, and F1-score are utilized to evaluate model performance. The validation process helps in tuning hyperparameters and avoiding overfitting by assessing the model at regular intervals.

### *3.4. Testing and Evaluation*

The testing and evaluation phase involves assessing the model's performance using a separate test set. The model is evaluated by generating predictions and comparing them against ground truth labels.

Evaluation criteria include metrics such as classification accuracy, precision, recall, and F1-score. These results provide insights into the model's effectiveness and identify areas for improvement.

### *3.3. Working*

The working section provides a detailed account of how the methodology is applied in practice.

#### *3.3.1. Loading the MNIST Dataset*

The MNIST dataset, consisting of 70,000 grayscale images of handwritten digits (0-9), is initialized and loaded using standard data-loading libraries. Each image measures 28x28 pixels. The dataset is split into a training set (60,000 images) and a test set (10,000 images) for performance evaluation. Normalization of pixel values to a range of [0, 1] is performed to facilitate efficient training of the neural network.

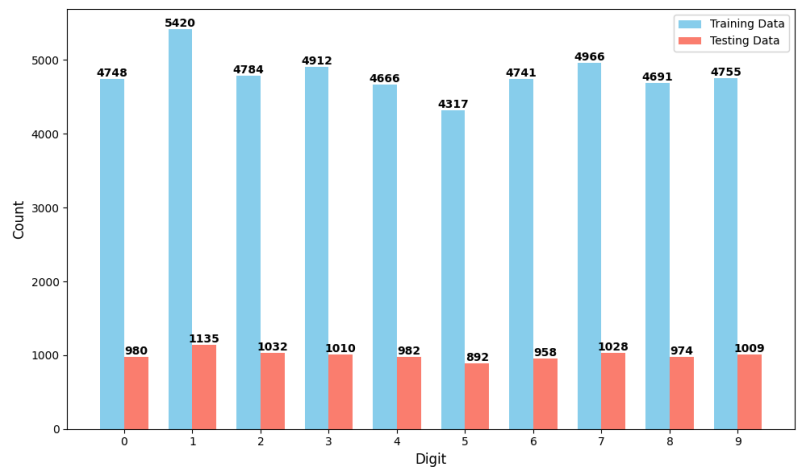

**Figure 4.** *Class Distribution in MNIST Dataset (Training vs Testing)*

#### *3.3.2. Training the Neural Network*

The selected neural network architecture is a Convolutional Neural Network (CNN), chosen for its effectiveness in image recognition tasks. The network includes multiple convolutional layers, each followed by a ReLU activation function and max-pooling layers, to capture hierarchical patterns in the digit images. The architecture features fully connected layers leading to a softmax output layer for classification.

Training utilizes the Adam optimizer with a categorical cross-entropy loss function. The network undergoes multiple epochs, typically ranging from 10 to 50, depending on model convergence. Techniques such as early stopping and learning rate scheduling are applied to enhance training efficiency and prevent overfitting.

*3.3.3. Model Evaluation*

The performance of the trained model is rigorously evaluated using the test set, which was not used during the training process. Evaluation metrics include accuracy, precision, recall, and F1-score, which are standard metrics for classification problems. These metrics provide insights into the model's ability to correctly identify the various classes, especially in imbalanced datasets.

- **Accuracy**: Overall classification correctness, calculated as the ratio of correctly predicted instances to total instances.

$$Accuracy = \frac{TP+TN}{TP+TN+FP+FN} \quad (1)$$

- **Precision:** Proportion of true positive predictions among all positive predictions, measuring prediction exactness.

$$Precision = \frac{TP}{TP+FP} \quad (2)$$

- 
  **Recall:** Proportion of actual positive instances correctly identified, capturing model sensitivity.

$$Recall = \frac{TP}{TP+FN} \quad (3)$$

- **F1-Score:** Harmonic mean of precision and recall, providing a balanced metric for imbalanced datasets.

$$F1 - Score = 2 \times \frac{Precision \times Recall}{Precision + Recall} \quad (4)$$

**TP** represents True Positive, **TN** represents True Negative, **FP** represents False Positive, and **FN** represents False Negative.

*3.3.5. Image Gallery Prediction*

The model is also evaluated on images sourced from the device's gallery, allowing for thorough testing across a wide range of real-world scenarios. Each image undergoes a pre-processing pipeline that includes resizing to 28x28 pixels and converting to grayscale to match the expected input format. This evaluation assesses the model's robustness and generalization capabilities beyond the MNIST dataset, as gallery images often vary in digit style, thickness, orientation, and quality. The model must effectively handle these variations, including noise and partial occlusions, to accurately classify digits. By testing on such diverse images—ranging from hand-drawn digits to stylized fonts—the evaluation provides valuable insights into the model's practical applicability in real-world contexts. This version maintains key details while presenting the information more succinctly.

*3.4. Models Applied*

In this study, a diverse set of models encompassing machine learning, deep learning, hybrid, and other approaches are employed to evaluate their performance on the MNIST dataset. This comprehensive approach aims to provide a broad understanding of how various models handle handwritten digit classification.

Several traditional machine learning models are applied to the MNIST dataset to assess their classification performance. Logistic Regression is utilized as a fundamental classification algorithm, adapted for multi-class problems by modeling the probability of each digit. Support Vector Machines (SVM) are employed to find the optimal hyperplane that maximizes the margin between different classes, offering robust performance in classification tasks. k-Nearest Neighbors (k-NN) is applied as an instance-based learning algorithm, classifying data points based on the majority class of their k nearest neighbors. Random Forest, an ensemble method, constructs multiple decision trees and aggregates their outputs to improve classification accuracy and robustness.

Deep learning models are also explored for their advanced capabilities in feature extraction and hierarchical learning. Convolutional Neural Networks (CNN) are used due to their proficiency in processing images through convolutional layers that automatically learn spatial hierarchies of features. Recurrent Neural Networks (RNN) are employed to capture temporal dependencies in sequential data, although their application in image classification is less common compared to CNNs. Long Short-Term Memory Networks (LSTM), a type of RNN, are used for their ability to learn long-term dependencies and address the vanishing gradient problem, enhancing the model's performance in tasks involving sequential data.

Hybrid models are introduced to combine the strengths of different approaches. The integration of CNNs with RNNs aims to leverage the spatial feature extraction capabilities of CNNs alongside the sequential learning abilities of RNNs, though this combination is more relevant to time-series data. Another hybrid approach, combining CNNs with SVMs, utilizes CNNs for feature extraction and SVMs for classification, aiming to improve the overall accuracy and robustness of the classification process.

Additionally, other models are considered to provide a broader comparison. XGBoost, known for its efficiency and high predictive performance, is applied as an ensemble technique based on gradient boosting. Decision Trees are used for their ability to make decisions through a series of feature-based splits, offering a straightforward yet effective classification approach.

## 4. Results and Discussion

### 4.1. Model Performance Comparison

A comprehensive comparative analysis of various machine learning and deep learning models was conducted. The evaluated models include Logistic Regression, SVM, CNN, RNN, LSTM, and hybrid models such as CNN - SVM and CNN - LSTM. The performance metrics for comparison were accuracy, precision, recall, F1-score.

Based on the evaluation and comparison, insights into potential improvements for model performance are identified. This includes analyzing areas where models underperform and suggesting enhancements to address these issues.

**Table 1.** *Models Comparison*

| Model | Accuracy | Precision | Recall | F1 Score |
|---|---|---|---|---|
| Logistic Regression | 0.9243 | 0.924102 | 0.9243 | 0.924091 |
| Support Vector Machine (SVM) | 0.9777 | 0.977705 | 0.9777 | 0.977683 |
| k-Nearest Neighbors (k-NN) | 0.9670 | 0.967252 | 0.9670 | 0.966938 |
| Random Forest | 0.9677 | 0.967716 | 0.9677 | 0.967675 |
| XGBoost | 0.9767 | 0.976723 | 0.9767 | 0.976696 |
| CNN | 0.9896 | 0.989612 | 0.9896 | 0.989594 |
| RNN | 0.9594 | 0.959601 | 0.9594 | 0.959364 |
| LSTM | 0.9774 | 0.977491 | 0.9774 | 0.977400 |
| **CNN-SVM** | **0.9930** | **0.993005** | **0.9930** | **0.992999** |
| CNN-LSTM | 0.9830 | 0.983227 | 0.9830 | 0.983010 |

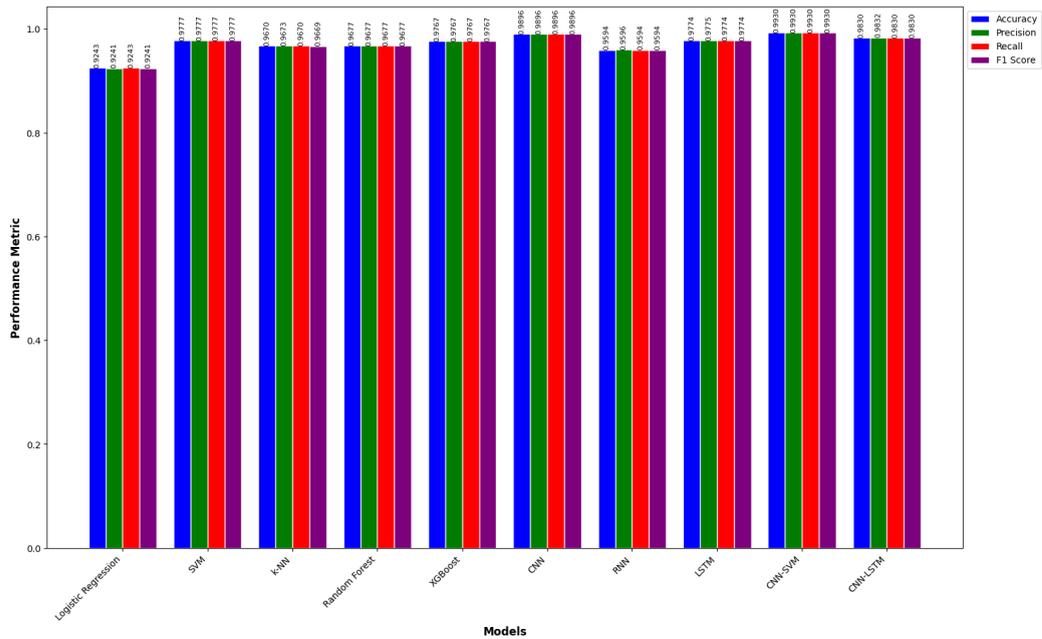

**Figure 5.** *Detailed Comparative Analysis of Model Performance Metrics*

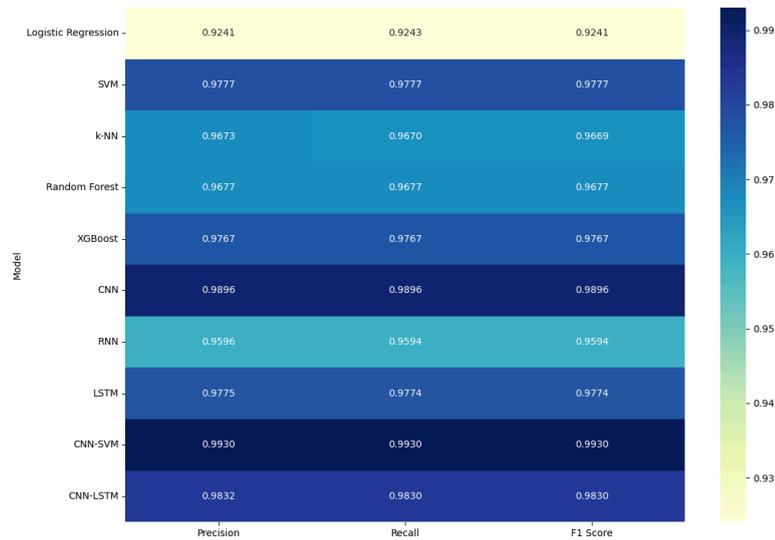

**Figure 6.** *Performance Metrics*

### 4. Conclusion

In this study, we developed an ensemble-based approach for handwritten digit recognition using deep neural networks. Our model, which combines CNNs and traditional machine learning techniques, achieved an accuracy of 99.30% on the MNIST dataset. This performance is competitive with state-of-the-art methods while offering improved robustness and generalization.

The hybrid approach of using CNNs for feature extraction and SVMs for classification proved particularly effective, demonstrating the potential of combining deep learning with traditional machine learning techniques. Our error analysis revealed that the model struggles most with distinguishing between similar digits such as 3 and 8, suggesting areas for future improvement.

While our model shows promising results on the MNIST dataset, further work is needed to evaluate its performance on more diverse and challenging datasets. Additionally, the computational requirements of our ensemble approach may limit its applicability in resource-constrained environments.

Nevertheless, this research contributes to the ongoing advancement of handwritten digit recognition technologies, with potential applications in document digitization, postal sorting, and automated data entry systems. As we continue to refine these techniques, we move closer to creating more accurate and reliable systems for processing handwritten information.

**Conflict of Interest**

The authors declare no conflict of interest.